\documentclass{article} 

\PassOptionsToPackage{numbers, compress}{natbib}
\usepackage[final]{neurips_2022}

\usepackage{hyperref}
\usepackage{url}
\usepackage{booktabs}
\usepackage{multicol}
\usepackage{graphicx}
\usepackage{xcolor}
\usepackage{colortbl} 
\usepackage{array}

\title{Contrastive Self-Supervised Learning \\ for Skeleton Representations}

\author{%
  Nico~Lingg\thanks{Work done during an internship at Apple.} \\
  Personal Robotics Lab \\ Imperial College London \\
  \texttt{n.lingg20@imperial.ac.uk} \\
  \And
  Miguel~Sarabia, Luca~Zappella, Barry-John~Theobald \\[0.18cm]
  Apple\\[0.18cm]
  \texttt{\{miguelsdc, lzappella, bjtheobald\}@apple.com} \\
}

\begin{document}

\maketitle

\begin{abstract}
Human skeleton point clouds are commonly used to automatically classify and predict the behaviour of others. In this paper, we use a contrastive self-supervised learning method, SimCLR, to learn representations that capture the semantics of skeleton point clouds. This work focuses on systematically evaluating the effects that different algorithmic decisions (including augmentations, dataset partitioning and backbone architecture) have on the learned skeleton representations. To pre-train the representations, we normalise six existing datasets to obtain more than 40 million skeleton frames. We evaluate the quality of the learned representations with three downstream tasks: skeleton reconstruction, motion prediction, and activity classification. Our results demonstrate the importance of 1) combining spatial and temporal augmentations, 2) including additional datasets for encoder training, and 3) and using a graph neural network as an encoder.
\end{abstract}

\section{Introduction}
\begin{figure}[t]
    \centering
    \includegraphics[width=\textwidth]{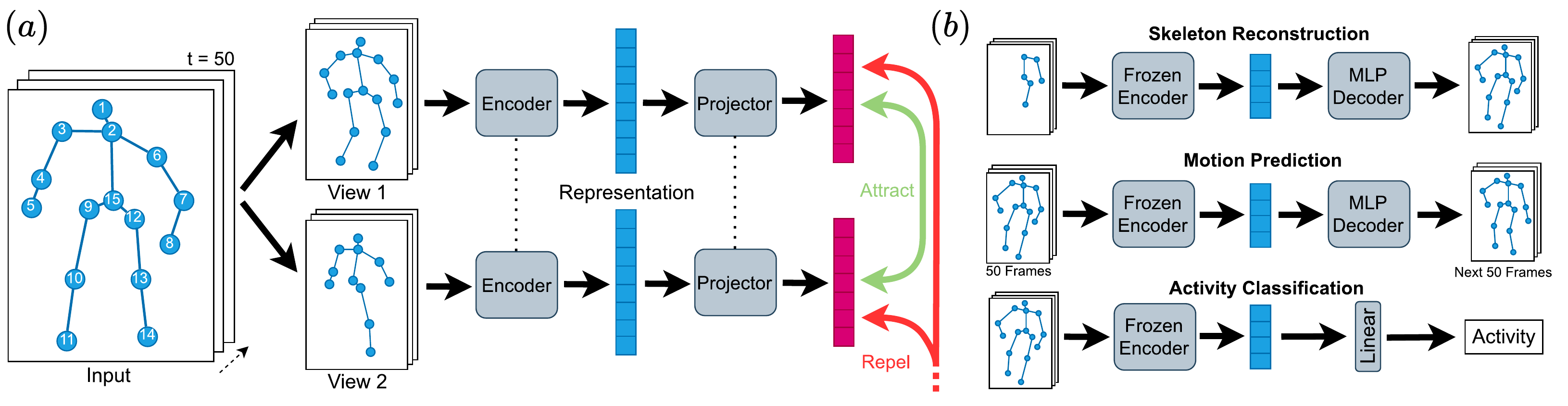}
    \caption{(a) Adapted architecture of SimCLR~\cite{SimCLR2020} for skeleton data. We use a custom augmentation stack (cf. Appendix~\ref{sec:augmentations}) and a ST-GCN~\cite{stgcn2018aaai} encoder. (b) Downstream tasks to evaluate the quality of the learned representation: skeleton reconstruction, motion prediction, and activity classification. Joints: 1) head, 2) neck, 3) right shoulder, 4) right elbow, 5) right wrist, 6) left shoulder, 7) left elbow, 8) left wrist, 9) right hip, 10) right knee, 11) right ankle, 12) left hip, 13) left knee, 14) left ankle, 15) torso.}
    \label{fig:simclr}
    \vspace{-5mm}
\end{figure}

Skeleton point clouds are commonly used to understand human behaviour, including activity classification~\cite{chen2021channel, stgcn2018aaai}, gait analysis~\cite{RashmiGait}, and command recognition~\cite{Fanello13a}; in fact, many sensors have been developed to easily capture skeleton data, whether through depth cameras or motion capture systems. However, working with skeletons has inherent challenges, such as occlusions, noisy data, missing joints, and poor skeleton fittings~\cite{Plantard2015}. Self-supervised learning has successfully been used to learn representations robust to noise and missing data in computer vision~\cite{KhanSSLSurvey}, speech recognition~\cite{Wang2022} and natural language processing~\cite{BrownNLP}. Inspired by this success, we aim to train a skeleton representation which generalises to three downstream tasks: skeleton reconstruction, motion prediction, and activity classification (see Fig.~\ref{fig:simclr}b).

We use SimCLR~\cite{SimCLR2020}, a multi-view contrastive self-supervised learning algorithm, which relies on augmentations to create two views of the same input sample. SimCLR uses a loss which attracts the representations of views from the same samples while repelling the representations of views from different samples (cf. Fig.~\ref{fig:simclr}a). We adapt SimCLR using a graph neural network backend, ST-GCN~\cite{stgcn2018aaai}, and employ eight augmentations (both spatial and temporal) to generate the views. It is worth noting that the choice of augmentations is a form of inductive bias -- it determines what invariances will be built into the learned representations. In this work, we seek representations that are invariant to movement speeds, the framerate of the recording, and sensor noise. 

Training with SimCLR requires making many algorithmic decisions involving network architecture, datasets, augmentations, etc. Our focus is to determine the effect of these decisions on downstream performance, hence we run exhaustive experimental ablations. Our results highlight the importance of combining both spatial and temporal augmentations, as well as the impact of using ST-GCN as an encoder with additional datasets to learn the representation. In fact, we gather six existing skeleton datasets captured with different sensors, subjects, conditions, and labels, which results in approximately 40 million frames, each with 15 3d joint positions.

\section{Background}
Our work is at the intersection of both multi-view self-supervised learning, and skeleton representation learning; therefore we will review the relevant background of both fields.

\textbf{Multi-View Self-Supervised Learning}. Multi-view self-supervised learning is a form of representation learning that uses augmentations to generate different views of the input. The views are then used to contrast between the representations of views from the same input and representations of views from other inputs. In recent years, various multi-view self-supervised learning frameworks have been proposed, including CPC~\cite{cpcdeepmind}, MoCO~v3~\cite{mocov3}, DINO~\cite{dino} and SimCLR~\cite{SimCLR2020}. These approaches have shown impressive results in transferability to smaller datasets like CIFAR 10, where DINO achieves an accuracy of 99.6\%. SimCLR, in particular, has been shown to be flexible, stable to train and does not need extra logic to mine negatives samples~\cite{SimCLR2020}. For these reasons, we make use of SimCLR in this work.

\textbf{Skeleton Representation Learning.} There are several related approaches to learning representations of skeleton point-clouds. In particular,~\cite{Su2021}~introduced a custom pretext task based on motion consistency and motion continuity as well as an ST-GCN encoder to learn skeleton representations. For this work, we reuse ST-GCN as our representation encoder. More recently,~\cite{chenTransformer}~made use of a transformer architecture and pretext tasks inspired by BERT~\cite{Devlin2019BERTPO} to achieve the state-of-the-art in activity classification accuracy. Additionally, CP-STN~\cite{pmlr-v157-zhan21a} combines contrastive learning with pretext tasks (e.g., masked sequences prediction) to extract spatial and temporal information to learn fully discriminative representations. Other works, like CrosSCLR~\cite{li2021crossclr}, further attempt to improve contrastive learning by using extra hard contrastive pairs stored in a memory bank. Unlike these approaches, which use custom pretext tasks to learn their representations, we take advantage of SimCLR to train the representation encoder.

\section{Learning a Skeleton Representation with SimCLR}
\label{sec:simclr}

We aggregate six skeleton datasets captured with different sensors, subjects, conditions, and labels (refer to Table~\ref{table:dataset_statitics} for an overview). We hypothesise that the combined diversity of the data leads to better representations (as shown in Section~\ref{sec:ablations}). Our custom pre-processing pipeline\footnote{Code available at: \href{https://github.com/apple/ml-skeleton-preprocessing}{https://github.com/apple/ml-skeleton-preprocessing}} consists of the following five steps: 1) transform the datasets joint representation into 3d global coordinates, 2) reduce the joint-space to 15 fundamental joints commonly provided by skeleton sensing systems, 3) scale each skeleton to a height of 2m, i.e. have the z-axis extend from -1.0 to 1.0, 4) translate the skeleton so that the torso is at the centre-of-coordinates, and 5) rotate each frame around the x- and z-axes so that the skeleton faces the camera. We do not alter frame-rate, nor the skeleton structure, so each skeleton retains its scaled bone lengths. 



\begin{table}[t]
\centering
\small
\caption{Datasets used in this work. Training set of datasets marked with $\ast$ are used to pre-train skeleton representation with SimCLR, those marked with $\diamond$ are used for downstream task training.}
\label{table:dataset_statitics}
\begin{scriptsize}
\begin{tabular}{%
    m{0.25\columnwidth}
    >{\raggedleft\arraybackslash}m{0.13\columnwidth}
    >{\raggedleft\arraybackslash}m{0.12\columnwidth}
    >{\raggedleft\arraybackslash}m{0.09\columnwidth}
    >{\centering\arraybackslash}m{0.11\columnwidth}
    >{\centering\arraybackslash}m{0.13\columnwidth}
}
\toprule

\textbf{\textsc{Dataset}} & %
\textbf{\textsc{Number of Train~Frames}} & %
\textbf{\textsc{Number of Test~Frames}} & %
\textbf{\textsc{Unique Human Subjects}} & %
\textbf{\textsc{Frames per Second}} & %
\textbf{\textsc{Recording Device}} \\
\midrule

\textsc{NTU RGB+D 60}$^{\ast\diamond}$~\cite{shahroudy2016ntu} & %
    3,328,560 (8\%) &       
    1,456,493 (92\%) &      
    40 (22\%)  &            
    30  &                   
    Depth Camera \\     
    
\textsc{NTU RGB+D 120}$^\ast$~\cite{liu2020ntu} & 
    4,324,668 (11\%) &      
    -- &                    
    66 (37\%) &             
    30 &                    
    Depth Camera \\     

\textsc{Trinity Speech-Gesture}$^\ast$~\cite{trinity} & 
    4,579,806  (11\%) &     
    -- &                    
    1 (1\%) &               
    120 &                   
    MoCap System \\             

\textsc{Talking with Hands}$^\ast$~\cite{talkingwithhands} & 
    25,680,106 (64\%) &     
    -- &                    
    50 (28\%) &             
    90 &                    
    MoCap System \\            

\textsc{DanceDB}$^\ast$~\cite{dancedb} & 
    1,679,209 (4\%) &       
    -- &                    
    17 (9\%) &              
    120 &                   
    MoCap System \\            

\textsc{HDM05}$^\diamond$~\cite{cg-2007-2}  & 
    492,340 (1\%) &                    
    123,085 (8\%) &        
    5 (3\%) &               
    120 &                   
    MoCap System \\             

\midrule
\textsc{Total} & 40,084,689 & 1,579,578 & 179  & -- \\
\bottomrule
\end{tabular}
\end{scriptsize}
\vspace{-1mm}
\end{table}

We pre-train the skeleton representation by feeding two augmented views of a 50-frame skeleton sequence into a graph encoder (ST-GCN~\cite{stgcn2018aaai}) which outputs a vector of 128 elements. The representation is then projected through a 3-layer multi-layer perceptron (MLP). Lastly, we use the NT-Xent loss~\cite{SimCLR2020} to attract projections from the same skeleton sequences and repel projections from different inputs. We pre-train SimCLR for 500 epochs with the datsets described above and then keep only the trained encoder for downstream tasks. We experimented with a larger number of epochs for pre-training, but found no improvement in downstream task performance. Pre-training and training hyperparameters are fixed across all experiments and are detailed in Appendix~\ref{sec:hyperparameters}.

To obtain the different views of our skeleton sequences, we use eight augmentations from two categories: spatial augmentations (joint dropout, joint jitter, axis mirroring, random scaling, frame dropout), and temporal augmentations (speed-up, slow-down). We describe the augmentations in detail in Appendix~\ref{sec:augmentations}. The purpose of these augmentations is to learn a skeleton representation invariant to noise in the skeleton joints and the actual speed of the human movements.

\section{Ablations on Skeleton Representations} \label{sec:ablations}

We analyse the effect of different algorithmic decisions on the quality of the learned representations by measuring the impact of these changes on the performance of the following downstream tasks:

\textbf{Skeleton Reconstruction}: The input is a sequence of skeleton frames from which 80\% of randomly selected frames have been dropped. The target is the uncorrupted skeleton sequence. We use a 3-layer MLP decoder, a mean-squared error (MSE) loss, and evaluate against the NTU-60 and HDM05 datasets. We report the test mean absolute joint error (MAE) in millimetres for a two-metre tall normalised skeleton. We introduce the MAE metric to have a consistent measure across test datasets. 

\textbf{Motion Prediction}: The target is to predict the next 50 frames of motion given the representation of the previous 50 frames of data. As above, we use the 3-layer MLP, MSE loss, evaluate on NTU-60 and HDM05, and report the test MAE in millimetres. We use MAE for consistency reasons (e.g. ~\cite{stgcn2018aaai} and~\cite{chenTransformer}).

\textbf{Activity Classification}: We use NTU-60~\cite{shahroudy2016ntu} labels, a linear classifier, and cross-entropy loss to classify motion segments. We use 50 frames from one skeleton to train the linear classifier rather than the standard 300 frames from two skeletons commonly used in the literature. Moroever, some of the activities in NTU-60 require joints not present in our representation, making it challenging to draw comparisons against existing approaches which take advantage of all the joints.

\newcommand{\separator}[1]{\midrule \multicolumn{6}{c}{\textsc{#1}}\\\midrule}

\begin{table}[t]
\centering
\caption{Effect of various ablations for training skeleton representations with SimCLR on downstream tasks. Each cell denotes mean performance and standard deviation across 3 different random seeds. For each row in the table, we pre-train 3 different encoder networks/linear classifiers using SimCLR and train each network on the downstream task with a frozen backbone. Baseline configuration has all augmentations (except random rotation), is trained with the full training set size, and was pre-trained with all five skeleton datasets. We highlighted changes over 8\% in bold.}
\label{table:ablations}
\begin{scriptsize}
\hspace*{-1.3cm}
\begin{tabular}{
    >{\raggedleft\arraybackslash}m{2.5cm}%
    >{\centering\arraybackslash}m{2.2cm}%
    @{\hspace{0.5cm}}%
    >{\centering\arraybackslash}m{2.2cm}%
    @{\hspace{0.5cm}}%
    >{\centering\arraybackslash}m{2.2cm}%
    @{\hspace{0.5cm}}%
    >{\centering\arraybackslash}m{2.2cm}%
    @{\hspace{0.5cm}}%
    >{\centering\arraybackslash}m{2.2cm}%
}
\toprule

& \multicolumn{2}{c}{\textbf{\textsc{Skeleton Reconstruction}}} & \multicolumn{2}{c}{\textbf{\textsc{Motion Prediction}}} & \textbf{\textsc{Activity Class.}} \\[0.1cm]
& \multicolumn{2}{c}{\textsc{Mean Absolute Error} (mm)} & \multicolumn{2}{c}{\textsc{Mean Absolute Error} (mm)} & \textsc{Accuracy} (\%) \\
& \multicolumn{2}{c}{\textit{Lower is better}} & \multicolumn{2}{c}{\textit{Lower is better}} & \textit{Higher is better} \\[0.1cm]
& \textsc{NTU-60} & \textsc{HDM05} & \textsc{NTU-60} & \textsc{HDM05} & \textsc{NTU-60}  \\

\midrule

\textsc{Baseline} & %
    \cellcolor[rgb]{0.9657, 0.9672, 0.9681} 62.0 ($\pm$ 2.3) &  
    \cellcolor[rgb]{0.9691, 0.9665, 0.9649} 49.7 ($\pm$ 1.5) &  
    \cellcolor[rgb]{0.9691, 0.9665, 0.9649} 52.1 ($\pm$ 0.6) &  
    \cellcolor[rgb]{0.9657, 0.9672, 0.9681} 23.4 ($\pm$ 1.0) &  
    \cellcolor[rgb]{0.9691, 0.9665, 0.9649} 45.8 ($\pm$ 1.0) \\ 

\textsc{Finetuning} & %
    \cellcolor[rgb]{0.8780, 0.9257, 0.9519} 65.7 ($\pm$ 1.7), $\Delta$+3.7 &  
    \cellcolor[rgb]{0.8002, 0.8882, 0.9356} 55.0 ($\pm$ 2.9), \textbf{$\Delta$+5.3} &  
    \cellcolor[rgb]{0.8430, 0.9091, 0.9455} 56.5 ($\pm$ 1.2), \textbf{$\Delta$+4.4} &  
    \cellcolor[rgb]{0.0196, 0.1882, 0.3804}\color{white} 44.7 ($\pm$ 3.2), \textbf{$\Delta$+21.3} &  
    \cellcolor[rgb]{0.6577, 0.0812, 0.1622}\color{white} 64.7 ($\pm$ 1.0), \textbf{$\Delta$+18.9} \\ 

\textsc{30 FPS} & %
    \cellcolor[rgb]{0.9728, 0.9493, 0.9354} 60.8 ($\pm$ 0.7), $\Delta$-1.2 &  
    \cellcolor[rgb]{0.8430, 0.9091, 0.9455} 53.8 ($\pm$ 1.6), \textbf{$\Delta$+4.1} &  
    \cellcolor[rgb]{0.9482, 0.9589, 0.9649} 52.8 ($\pm$ 0.6), $\Delta$+0.7 &  
    \cellcolor[rgb]{0.0196, 0.1882, 0.3804}\color{white} 37.1 ($\pm$ 0.6), \textbf{$\Delta$+13.7} &  
    \cellcolor[rgb]{0.9482, 0.9589, 0.9649} 45.1 ($\pm$ 0.1), $\Delta$-0.7 \\ 

\textsc{MLP Backbone} & %
    \cellcolor[rgb]{0.9073, 0.9396, 0.9573} 64.6 ($\pm$ 2.3), $\Delta$+2.6 &  
    \cellcolor[rgb]{0.6161, 0.7947, 0.8830} 58.8 ($\pm$ 3.3), \textbf{$\Delta$+9.1} &  
    \cellcolor[rgb]{0.7421, 0.8587, 0.9190} 59.0 ($\pm$ 2.3), \textbf{$\Delta$+6.9} &  
    \cellcolor[rgb]{0.3599, 0.6380, 0.7979}\color{white} 29.7 ($\pm$ 1.5), \textbf{$\Delta$+6.3} &  
    \cellcolor[rgb]{0.1791, 0.4657, 0.7081}\color{white} 29.3 ($\pm$ 0.9), \textbf{$\Delta$-16.5} \\ 

\separator{Downstream Task Training Set Size}
\textsc{15\%} & %
    \cellcolor[rgb]{0.9783, 0.9234, 0.8911} 59.5 ($\pm$ 0.9), $\Delta$-2.5 &  
    \cellcolor[rgb]{0.9014, 0.9368, 0.9562} 52.0 ($\pm$ 0.8), $\Delta$+2.3 &  
    \cellcolor[rgb]{0.7712, 0.8734, 0.9273} 58.3 ($\pm$ 1.0), \textbf{$\Delta$+6.2} &  
    \cellcolor[rgb]{0.9482, 0.9589, 0.9649} 23.7 ($\pm$ 0.9), $\Delta$+0.3 &  
    \cellcolor[rgb]{0.7033, 0.8390, 0.9080} 39.1 ($\pm$ 1.7), \textbf{$\Delta$-6.7} \\ 

\textsc{10\%} & %
    \cellcolor[rgb]{0.9829, 0.9019, 0.8542} 58.2 ($\pm$ 0.6), $\Delta$-3.8 &  
    \cellcolor[rgb]{0.8839, 0.9285, 0.9530} 52.6 ($\pm$ 1.2), $\Delta$+2.9 &  
    \cellcolor[rgb]{0.7227, 0.8488, 0.9135} 59.4 ($\pm$ 0.7), \textbf{$\Delta$+7.3} &  
    \cellcolor[rgb]{0.0196, 0.1882, 0.3804}\color{white} 54.3 ($\pm$ 2.3), \textbf{$\Delta$+30.9} &  
    \cellcolor[rgb]{0.5968, 0.7849, 0.8775} 37.1 ($\pm$ 1.0), \textbf{$\Delta$-8.7} \\ 

\textsc{5\%} & %
    \cellcolor[rgb]{0.9839, 0.8976, 0.8468} 57.9 ($\pm$ 0.5), $\Delta$-4.1 &  
    \cellcolor[rgb]{0.0498, 0.2464, 0.4611}\color{white} 73.1 ($\pm$ 1.3), \textbf{$\Delta$+23.4} &  
    \cellcolor[rgb]{0.0498, 0.2464, 0.4611}\color{white} 76.7 ($\pm$ 4.0), \textbf{$\Delta$+24.6} &  
    \cellcolor[rgb]{0.0196, 0.1882, 0.3804}\color{white} 73.5 ($\pm$ 6.6), \textbf{$\Delta$+50.1} &  
    \cellcolor[rgb]{0.3478, 0.6303, 0.7938}\color{white} 33.3 ($\pm$ 0.4), \textbf{$\Delta$-12.5} \\ 

\separator{SimCLR Pre-training datasets}
\textsc{w/o Trinity} & %
    \cellcolor[rgb]{0.9746, 0.9406, 0.9206} 60.4 ($\pm$ 0.4), $\Delta$-1.6 &  
    \cellcolor[rgb]{0.9691, 0.9665, 0.9649} 49.6 ($\pm$ 0.9), $\Delta$-0.1 &  
    \cellcolor[rgb]{0.9691, 0.9665, 0.9649} 51.9 ($\pm$ 0.6), $\Delta$-0.2 &  
    \cellcolor[rgb]{0.9540, 0.9617, 0.9659} 23.6 ($\pm$ 0.8), $\Delta$+0.2 &  
    \cellcolor[rgb]{0.9248, 0.9479, 0.9606} 44.5 ($\pm$ 1.5), $\Delta$-1.3 \\ 

\textsc{w/o TWH} & %
    \cellcolor[rgb]{0.9765, 0.9320, 0.9059} 60.0 ($\pm$ 0.2), $\Delta$-2.0 &  
    \cellcolor[rgb]{0.9482, 0.9589, 0.9649} 50.3 ($\pm$ 0.7), $\Delta$+0.6 &  
    \cellcolor[rgb]{0.9599, 0.9645, 0.9670} 52.4 ($\pm$ 0.4), $\Delta$+0.3 &  
    \cellcolor[rgb]{0.9700, 0.9622, 0.9576} 23.3 ($\pm$ 0.4), $\Delta$-0.1 &  
    \cellcolor[rgb]{0.9365, 0.9534, 0.9627} 44.9 ($\pm$ 0.5), $\Delta$-0.9 \\ 

\textsc{w/o DanceDB} & %
    \cellcolor[rgb]{0.9709, 0.9579, 0.9502} 61.3 ($\pm$ 0.6), $\Delta$-0.7 &  
    \cellcolor[rgb]{0.9248, 0.9479, 0.9606} 51.2 ($\pm$ 0.8), $\Delta$+1.5 &  
    \cellcolor[rgb]{0.9131, 0.9423, 0.9584} 54.0 ($\pm$ 1.7), $\Delta$+1.9 &  
    \cellcolor[rgb]{0.8956, 0.9340, 0.9552} 24.5 ($\pm$ 0.8), $\Delta$+1.1 &  
    \cellcolor[rgb]{0.9599, 0.9645, 0.9670} 45.5 ($\pm$ 1.4), $\Delta$-0.3 \\ 

\textsc{w/o NTU-60 \& 120} & %
    \cellcolor[rgb]{0.9131, 0.9423, 0.9584} 64.2 ($\pm$ 0.3), $\Delta$+2.2 &  
    \cellcolor[rgb]{0.9306, 0.9506, 0.9616} 50.9 ($\pm$ 0.6), $\Delta$+1.2 &  
    \cellcolor[rgb]{0.8897, 0.9313, 0.9541} 54.9 ($\pm$ 0.5), $\Delta$+2.8 &  
    \cellcolor[rgb]{0.8839, 0.9285, 0.9530} 24.7 ($\pm$ 0.5), $\Delta$+1.3 &  
    \cellcolor[rgb]{0.7809, 0.8784, 0.9301} 40.5 ($\pm$ 0.2), \textbf{$\Delta$-5.3} \\ 

\textsc{NTU-60 Only} & %
    \cellcolor[rgb]{0.9599, 0.9645, 0.9670} 62.4 ($\pm$ 0.7), $\Delta$+0.4 &  
    \cellcolor[rgb]{0.8605, 0.9174, 0.9487} 53.2 ($\pm$ 0.6), $\Delta$+3.5 &  
    \cellcolor[rgb]{0.9365, 0.9534, 0.9627} 53.3 ($\pm$ 0.3), $\Delta$+1.2 &  
    \cellcolor[rgb]{0.7518, 0.8636, 0.9218} 26.4 ($\pm$ 0.9), \textbf{$\Delta$+3.0} &  
    \cellcolor[rgb]{0.7130, 0.8439, 0.9107} 39.2 ($\pm$ 0.2), \textbf{$\Delta$-6.6} \\ 

\separator{Augmentations}
\textsc{w/o Axis Mirroring} & %
    \cellcolor[rgb]{0.9774, 0.9277, 0.8985} 59.8 ($\pm$ 1.2), $\Delta$-2.2 &  
    \cellcolor[rgb]{0.9755, 0.9363, 0.9133} 48.2 ($\pm$ 1.0), $\Delta$-1.5 &  
    \cellcolor[rgb]{0.9709, 0.9579, 0.9502} 51.5 ($\pm$ 0.9), $\Delta$-0.6 &  
    \cellcolor[rgb]{0.9737, 0.9449, 0.9280} 22.9 ($\pm$ 0.4), $\Delta$-0.5 &  
    \cellcolor[rgb]{0.9248, 0.9479, 0.9606} 44.5 ($\pm$ 0.5), $\Delta$-1.3 \\ 

\textsc{w/o Random Scaling} & %
    \cellcolor[rgb]{0.9599, 0.9645, 0.9670} 62.3 ($\pm$ 0.5), $\Delta$+0.3 &  
    \cellcolor[rgb]{0.9073, 0.9396, 0.9573} 51.8 ($\pm$ 0.5), $\Delta$+2.1 &  
    \cellcolor[rgb]{0.9190, 0.9451, 0.9595} 53.8 ($\pm$ 0.2), $\Delta$+1.7 &  
    \cellcolor[rgb]{0.6937, 0.8341, 0.9052} 26.9 ($\pm$ 0.3), \textbf{$\Delta$+3.5} &  
    \cellcolor[rgb]{0.5665, 0.7687, 0.8685} 36.5 ($\pm$ 0.5), \textbf{$\Delta$-9.3} \\ 

\textsc{w/o Joint Jitter} & %
    \cellcolor[rgb]{0.9657, 0.9672, 0.9681} 62.1 ($\pm$ 1.1), $\Delta$+0.1 &  
    \cellcolor[rgb]{0.9190, 0.9451, 0.9595} 51.4 ($\pm$ 0.7), $\Delta$+1.7 &  
    \cellcolor[rgb]{0.9482, 0.9589, 0.9649} 52.9 ($\pm$ 1.1), $\Delta$+0.8 &  
    \cellcolor[rgb]{0.9248, 0.9479, 0.9606} 24.1 ($\pm$ 0.9), $\Delta$+0.7 &  
    \cellcolor[rgb]{0.8897, 0.9313, 0.9541} 43.4 ($\pm$ 0.9), $\Delta$-2.4 \\ 

\textsc{w/o Slow Down} & %
    \cellcolor[rgb]{0.9691, 0.9665, 0.9649} 61.9 ($\pm$ 0.4), $\Delta$-0.1 &  
    \cellcolor[rgb]{0.9131, 0.9423, 0.9584} 51.5 ($\pm$ 1.3), $\Delta$+1.8 &  
    \cellcolor[rgb]{0.9365, 0.9534, 0.9627} 53.3 ($\pm$ 0.5), $\Delta$+1.2 &  
    \cellcolor[rgb]{0.8255, 0.9008, 0.9423} 25.6 ($\pm$ 1.5), \textbf{$\Delta$+2.2} &  
    \cellcolor[rgb]{0.8897, 0.9313, 0.9541} 43.3 ($\pm$ 0.6), $\Delta$-2.5 \\ 

\textsc{w/o Speed Up} & %
    \cellcolor[rgb]{0.9728, 0.9493, 0.9354} 60.9 ($\pm$ 0.5), $\Delta$-1.1 &  
    \cellcolor[rgb]{0.9540, 0.9617, 0.9659} 50.2 ($\pm$ 0.6), $\Delta$+0.5 &  
    \cellcolor[rgb]{0.9657, 0.9672, 0.9681} 52.2 ($\pm$ 0.4), $\Delta$+0.1 &  
    \cellcolor[rgb]{0.9599, 0.9645, 0.9670} 23.5 ($\pm$ 0.7), $\Delta$+0.1 &  
    \cellcolor[rgb]{0.9131, 0.9423, 0.9584} 44.1 ($\pm$ 1.5), $\Delta$-1.7 \\ 

\textsc{w/o Frame Dropout} & %
    \cellcolor[rgb]{0.9737, 0.9449, 0.9280} 60.6 ($\pm$ 0.6), $\Delta$-1.4 &  
    \cellcolor[rgb]{0.9691, 0.9665, 0.9649} 49.6 ($\pm$ 0.6), $\Delta$-0.1 &  
    \cellcolor[rgb]{0.9691, 0.9665, 0.9649} 51.9 ($\pm$ 1.4), $\Delta$-0.2 &  
    \cellcolor[rgb]{0.9482, 0.9589, 0.9649} 23.7 ($\pm$ 0.8), $\Delta$+0.3 &  
    \cellcolor[rgb]{0.9365, 0.9534, 0.9627} 44.9 ($\pm$ 0.7), $\Delta$-0.9 \\ 

\textsc{w/o Joint Dropout} & %
    \cellcolor[rgb]{0.7905, 0.8833, 0.9329} 68.8 ($\pm$ 1.4), \textbf{$\Delta$+6.8} &  
    \cellcolor[rgb]{0.7130, 0.8439, 0.9107} 56.7 ($\pm$ 2.6), \textbf{$\Delta$+7.0} &  
    \cellcolor[rgb]{0.9700, 0.9622, 0.9576} 51.8 ($\pm$ 1.1), $\Delta$-0.3 &  
    \cellcolor[rgb]{0.8488, 0.9119, 0.9466} 25.3 ($\pm$ 0.8), \textbf{$\Delta$+1.9} &  
    \cellcolor[rgb]{0.8488, 0.9119, 0.9466} 42.1 ($\pm$ 0.9), \textbf{$\Delta$-3.7} \\ 

\textsc{w/ Random Rotation} & %
    \cellcolor[rgb]{0.9423, 0.9562, 0.9638} 63.2 ($\pm$ 0.6), $\Delta$+1.2 &  
    \cellcolor[rgb]{0.8780, 0.9257, 0.9519} 52.7 ($\pm$ 1.2), $\Delta$+3.0 &  
    \cellcolor[rgb]{0.9190, 0.9451, 0.9595} 53.9 ($\pm$ 0.8), $\Delta$+1.8 &  
    \cellcolor[rgb]{0.8722, 0.9230, 0.9509} 24.9 ($\pm$ 0.7), $\Delta$+1.5 &  
    \cellcolor[rgb]{0.9802, 0.9148, 0.8764} 48.1 ($\pm$ 0.3), $\Delta$+2.3 \\ 

\textsc{Spatial Only} & %
    \cellcolor[rgb]{0.9540, 0.9617, 0.9659} 62.5 ($\pm$ 0.9), $\Delta$+0.5 &  
    \cellcolor[rgb]{0.9131, 0.9423, 0.9584} 51.5 ($\pm$ 1.7), $\Delta$+1.8 &  
    \cellcolor[rgb]{0.9073, 0.9396, 0.9573} 54.2 ($\pm$ 2.0), $\Delta$+2.1 &  
    \cellcolor[rgb]{0.8547, 0.9146, 0.9476} 25.2 ($\pm$ 0.8), $\Delta$+1.8 &  
    \cellcolor[rgb]{0.8430, 0.9091, 0.9455} 42.0 ($\pm$ 0.9), \textbf{$\Delta$-3.8} \\ 

\textsc{Temporal Only} & %
    \cellcolor[rgb]{0.6840, 0.8291, 0.9024} 71.6 ($\pm$ 0.2), \textbf{$\Delta$+9.6} &  
    \cellcolor[rgb]{0.6161, 0.7947, 0.8830} 58.8 ($\pm$ 3.4), \textbf{$\Delta$+9.1} &  
    \cellcolor[rgb]{0.9482, 0.9589, 0.9649} 52.8 ($\pm$ 0.3), $\Delta$+0.7 &  
    \cellcolor[rgb]{0.5300, 0.7456, 0.8561} 28.4 ($\pm$ 2.0), \textbf{$\Delta$+5.0} &  
    \cellcolor[rgb]{0.0196, 0.1882, 0.3804}\color{white} 23.0 ($\pm$ 0.1), \textbf{$\Delta$-22.8} \\ 


\bottomrule 
\end{tabular}
\end{scriptsize}
\vspace{-1mm}
\end{table}

The results of our ablations are presented in Table~\ref{table:ablations}. Note that we report the performance from the final epoch. Our baseline models, which consist of a linear classifier/decoder trained for 200 epochs on top of the frozen representation, achieve an MAE of between 50mm and 60mm in most skeleton reconstruction and motion prediction tasks (except in motion prediction for HDM05 where the MAE is 23.4mm) and 45.8\% accuracy in the 60-class activity classifier task~\footnote{For context,~\cite{stgcn2018aaai} and~\cite{chenTransformer} achieve 81.5\% and 90\% accuracy respectively. Still, a direct comparison cannot be made due to the reduced joint space and temporal receptive field of our models}.

We remark that while fine-tuning the representation yields an improvement in activity classification, it actually hurts performance on skeleton reconstruction and motion prediction. This is due to the training sets needing more samples to fully fine-tune the downstream tasks models. Our results also show that using a normalised frame-rate across all datasets of 30 fps does not improve performance. Replacing the ST-GCN backbone with an MLP decreases the performance across most tasks, most notably with a drop of 16.5\% in activity classification. 

The results confirm that task performance improves with the number of downstream training samples. Interestingly, we observed a 6.6\% decrease in accuracy when pre-training only with NTU-60. This shows the importance of high diversity in pre-training data. However, removing a single pre-training dataset does not have a sizeable effect on downstream performance. Likewise, removing individual augmentations does not have a particular effect, except for random scaling (e.g. the performance drops by 9.3\% for activity classification) and joint dropout (e.g. the MAE on skeleton reconstruction increases by 6.8mm for NTU-60 and 7mm for HDM05). However, when we remove \textit{all} spatial augmentations or \textit{all} temporal augmentations simultaneously, we observe a drop in performance across most tasks (e.g. with only temporal augmentations, the MAE on skeleton reconstruction increases by 9.6mm on NTU-60 and decreases by 22.8\% in accuracy for activity classification).

\section{Conclusion and Further Work}

In this work, we showed that both spatial and temporal augmentations are needed to train skeleton representations. However, the specific type of augmentation within each category is less critical. Our results indicate that more data can help the model learn generic representations, even if the dataset were collected for different purposes. Finally, we demonstrated that a graph-based encoder (ST-GCN ~\cite{stgcn2018aaai}) produces higher quality representations.

In the future, we would like to explore how other modalities, such as audio, can further enrich skeleton representations.

\newpage

\begin{small}
\subsubsection*{Acknowledgments}
We would like to thank Zak Aldeneh, Jason Ramapuram, Arno Blass, Xavier Suau, Federico Danieli, Andy Keller, Borja Rodríguez, Katherine Metcalf, Nick Apostoloff, Russ Webb, Josh Susskind, and the anonymous reviewers for their helpful and knowledgeable feedback on this article. Additionally, we are grateful to Jason Ramapuram and Dan Busbridge for their excellent scalable implementation of SimCLR.
\end{small}

\bibliography{bibliography}
\bibliographystyle{iclr2023_conference}

\newpage
\appendix
\section{Augmentations}
\label{sec:augmentations}
Our version of SimCLR for skeleton representations makes use of the following augmentations:

\begin{description}
\item[Axis mirroring:] Multiply by -1 all coordinates in the x, y or z axis.
\item[Random scale:] Randomly scale the dimension of a joint for all frames in the input.
\item[Joint Jitter:] Add gaussian noise (with mean zero and variance 0.02) to each joint and frame.
\item[Slow down:] Linearly interpolate between two frames and select random continuous subset.
\item[Speed up:] Skip 1 or 2 frames when preparing input representations. Pad with zeros.
\item[Frame Dropout:] Completely zeroing a frame on the input.
\item[Joint Dropout:] Randomly zero some joints from the input.
\item[Random rotation:] Randomly rotate the skeleton in XYZ for all frames in the input.
\end{description}

\begin{figure}[h]
    \centering
    \includegraphics[width=\textwidth]{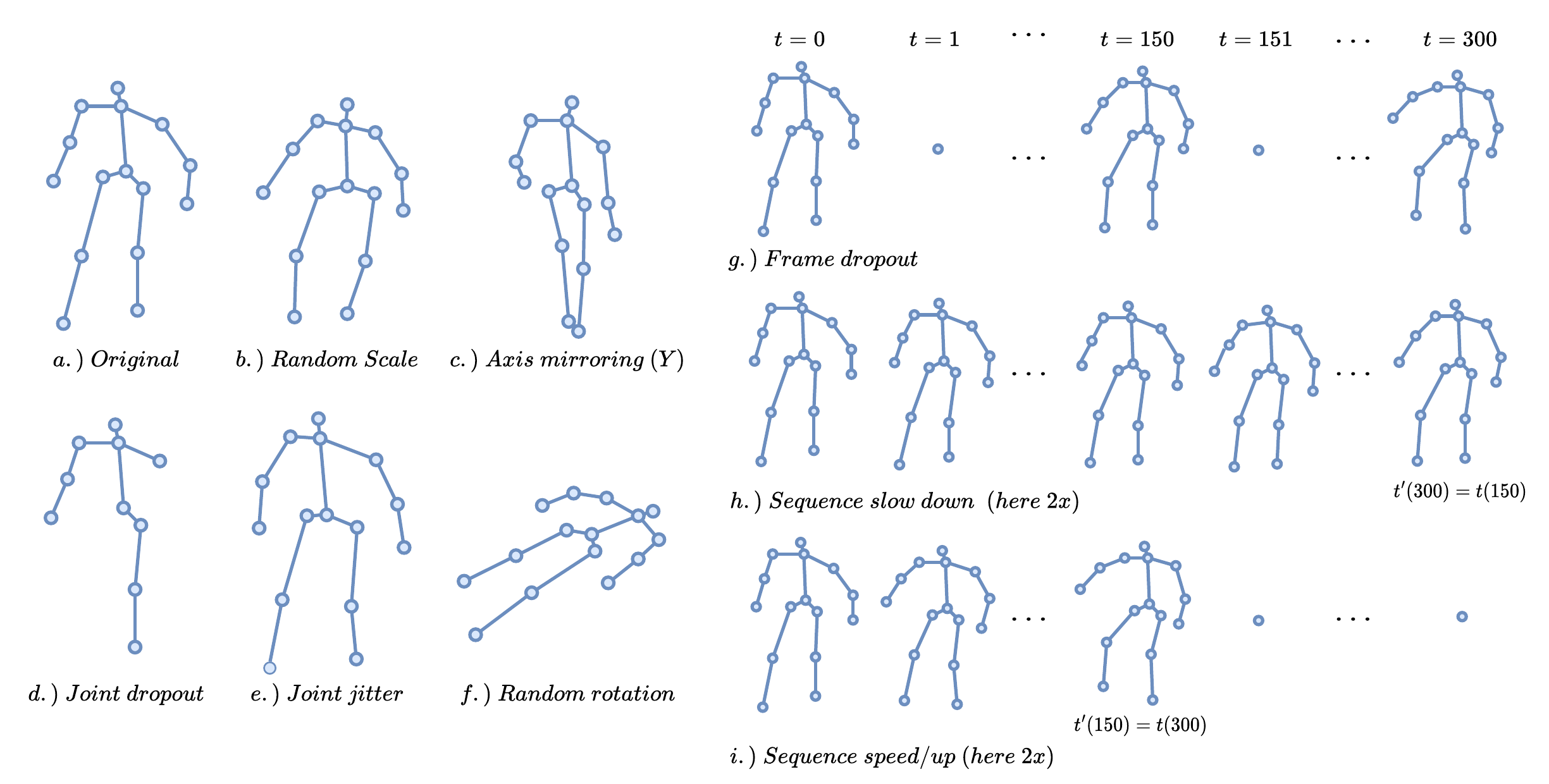}
    \caption{The augmentations used to train the SSL model with SimCLR. We use spatial (fig. b-g) and temporal(fig. h-i) augmentations. }
    \label{fig:augmentations}
\end{figure}

Visualisations of the augmentations can be seen in figure \ref{fig:augmentations}. The augmentations are applied to the input with the probabilities and strengths specified in table \ref{table:hyperparameters}. They are always applied in the order specified above.

\section{Training hyper-parameters}
\label{sec:hyperparameters}

\begin{table}[h]
\centering
\caption{Augmentations hyper-parameters for pre-training with probabilities and strengths}
\label{table:hyperparameters}
\begin{scriptsize}
\normalsize
\begin{tabular}{%
       >{\centering\arraybackslash}m{2.5cm}%
       >{\centering\arraybackslash}m{2.5cm}%
       >{\centering\arraybackslash}m{2.5cm}%
}
\toprule

Augmentation    &  Probability  &  Strength                     \\

\midrule

Axis mirroring  &  0.5          &  -                            \\ 
Random Scale    &  0.7          &  $\mu$: 0, $\sigma^2$: 0.02   \\
Joint Jitter    &  0.5          &  $\mu$: 0, $\sigma^2$: 0.02   \\
Slow down       &  0.5          &  0.5x-0.9x                    \\ 
Speed up        &  0.5          &  1.2-2x                       \\ 
Frame Dropout   &  0.5          &  0.5                          \\ 
Joint Dropout   &  0.5          &  0.5                          \\ 
Random rotation &  0.6          &  $\pm \pi$                    \\

\bottomrule
\end{tabular}
\end{scriptsize}
\end{table}

\begin{table}[t!]
\centering
\caption{Hyper-parameter used in pretraining and downstream tasks}
\begin{scriptsize}
\normalsize
\begin{tabular}{%
      >{\centering\arraybackslash}m{3.5cm}%
      >{\centering\arraybackslash}m{3.5cm}%
      >{\centering\arraybackslash}m{3.5cm}%
}

\toprule
& Pretraining & Downstream Task\\
\midrule
Batch Size              & 8096           & 1024  \\
Training Epochs         & 500            & 200   \\
Optimizer               & Lars Momentum  & Adam  \\
Learning Rate           & 0.01           & 0.01  \\
LR Warmup               & 10             & -     \\
LR Update Schedule      & cosine         & -     \\
Backbone Feature Size   & 128   & -              \\
Head Latent Size        & 256   & -              \\
Head Output Size        & 128   & -              \\
Head Output Size        & 128   & -              \\

\bottomrule
\end{tabular}
\end{scriptsize}
\end{table}

\begin{table}[t!]
\centering
\caption{Network Parameter used in pretraining and downstream tasks}
\begin{scriptsize}
\normalsize
\begin{tabular}{%
      >{\centering\arraybackslash}m{5cm}%
      >{\centering\arraybackslash}m{3.5cm}%
}
\toprule
& Trainable Parameters                          \\
\midrule
SimCLR &                                        \\
Encoder (ST-GCN)             & 0.968458M        \\
Encoder (MLP)                & 1.351168M        \\[0.1cm]
Frozen &                                        \\
Activity Classification         & 0.00774M      \\
Motion Prediction Frozen        & 2.905182M     \\
Skeleton Reconstruction Frozen  & 2.905182M     \\[0.1cm]
Fine-Tune &                                     \\
Activity Classification         & 0.843206M     \\
Motion Prediction Frozen        & 3.740648M     \\
Skeleton Reconstruction Frozen  & 3.740648M     \\
\bottomrule
\end{tabular}
\end{scriptsize}
\vspace{10cm}
\end{table}

\end{document}